\documentclass[final]{cvpr}

\usepackage{times}
\usepackage{epsfig}
\usepackage{graphicx}
\usepackage{subfigure}
\usepackage{amsmath}
\usepackage{amsthm}
\usepackage{amssymb}
\usepackage{algorithm}
\usepackage{algorithmicx}
\usepackage[noend]{algpseudocode}
\usepackage{multirow}
\usepackage{bm}
\usepackage{float}
\usepackage{color}

\DeclareMathOperator{\tr}{tr}

\DeclareMathOperator{\rank}{rank}
\DeclareMathOperator{\blockdiag}{blockdiag}
\DeclareMathOperator{\symblockdiag}{symblockdiag}

\usepackage[pagebackref=true,breaklinks=true,colorlinks,bookmarks=false]{hyperref}

\def\confYear{CVPR 2021}

\begin{document}

\title{Hybrid Rotation Averaging: A Fast and Robust Rotation Averaging Approach}

\author{Yu Chen \\
Peking University\\
{\tt\small dreamerhacker34@gmail.com}
\and
Ji Zhao\\
TuSimple\\
{\tt\small zhaoji84@gmail.com}
\and
Laurent Kneip\\
ShanghaiTech University\\
{\tt\small lkneip@shanghaitech.edu.cn}
}

\maketitle

\pagestyle{empty}
\thispagestyle{empty}

\begin{abstract}

We address rotation averaging (RA) and its application to real-world 3D reconstruction. 
Local optimisation based approaches are the de facto choice, though they only guarantee a local optimum.
Global optimisers ensure global optimality in low noise conditions, but they are inefficient and may
easily deviate under the influence of outliers or elevated noise levels.
We push the envelope of rotation averaging by leveraging the advantages of a global RA method and a local RA method.
Combined with a fast view graph filtering as preprocessing, the proposed hybrid approach is robust to outliers.
We further apply the proposed hybrid rotation averaging approach to incremental Structure from Motion (SfM), 
the accuracy and robustness of SfM are both improved by adding the resulting global rotations as regularisers 
to bundle adjustment. Overall, we demonstrate high practicality of the proposed method as bad camera poses are 
effectively corrected and drift is reduced.

\end{abstract}

\section{Introduction}

Rotation averaging is a problem that consists of estimating absolute camera orientations that agree as well as possible with a set of pairwise relative orientations. Errors expressing 
disagreements between estimated absolute orientations and the measured relative orientations are hereby distributed 
over each pairwise constraint. Rotation averaging is essential in global or hierarchical Structure from Motion (SfM) 
\cite{DBLP:conf/iccv/MoulonMM13,DBLP:conf/iccv/CuiT15,DBLP:conf/iccv/SweeneySHTP15,zhu2017parallel,DBLP:conf/cvpr/ZhuZZSFTQ18},
 as well as Simultaneous Localization and Mapping (SLAM) \cite{DBLP:conf/icra/BustosCER19} where it can accelerate camera 
 pose estimation and reduce drift accumulation.

In global SfM, we typically start by constructing a view graph $\mathcal{G}$ that encodes all connections between pairs of 
views $i$ and $j$ by an edge $(i, j)$, each one including the relative motion between image
$i$ and image $j$. Rotation averaging then gives us the absolute orientation of each view, and it is typically followed 
by a translation averaging step \cite{DBLP:conf/iccv/JiangCT13,DBLP:conf/cvpr/OzyesilS15,DBLP:conf/eccv/GoldsteinHLVS16,
DBLP:journals/cvpr/ZhuangCL18} to also obtain absolute positions. Triangulation of 3D points and joint optimisation over 
all parameters (i.e. bundle adjustment~\cite{DBLP:conf/iccvw/TriggsMHF99}) completes the reconstruction. In SLAM, rotation 
averaging has been used in the back-end pose graph optimisation \cite{rosen2019se,DBLP:conf/icra/BustosCER19} to flexibly 
encounter large drift accumulations or---more generally---replace the time-consuming bundle adjustment step.

\begin{figure}[t]
    \centering
     \includegraphics[width=0.95\linewidth]{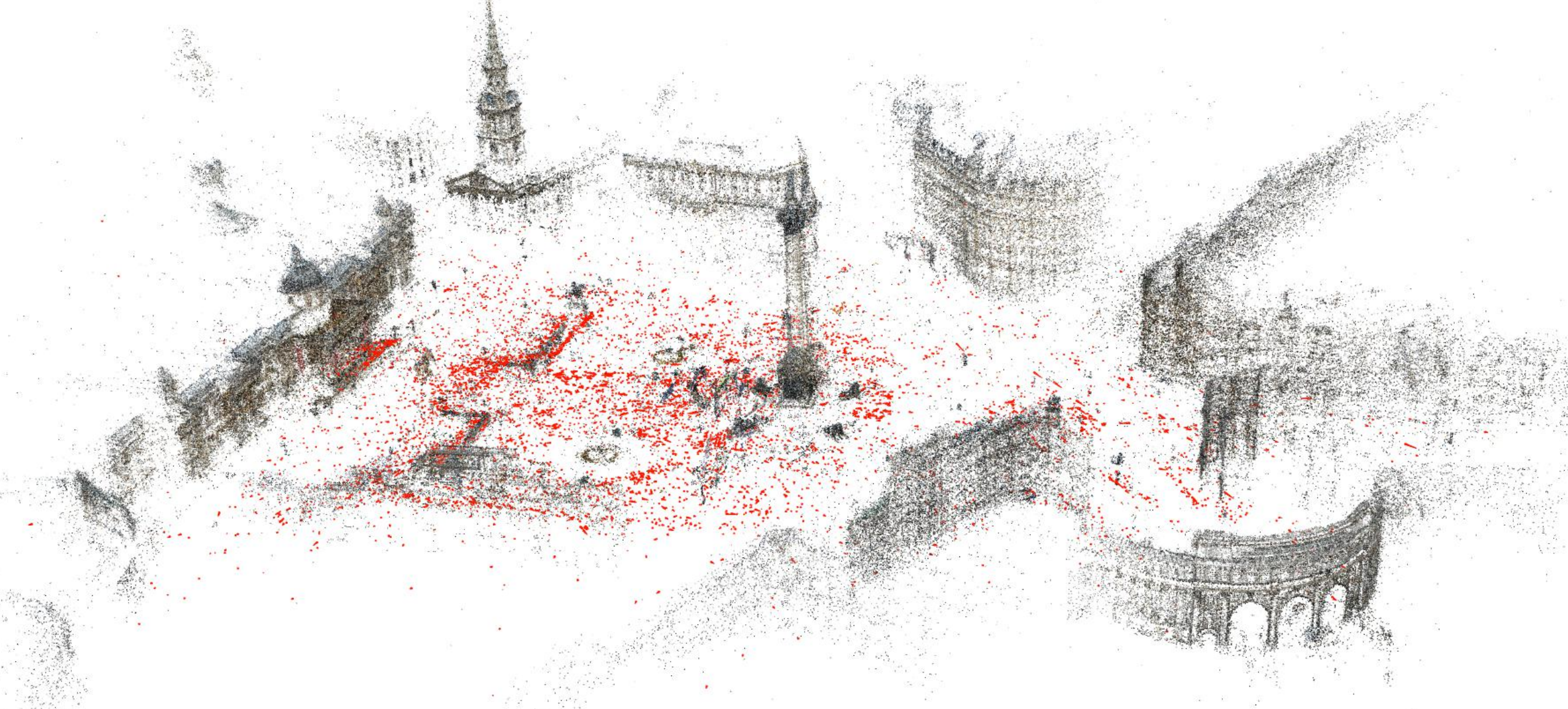}
    \caption{Reconstructions generated from the Trafalgar dataset~\cite{DBLP:conf/eccv/WilsonS14},
     where 7085 out of 15685 images have been registered, and the rotation averaging step only took 4.2 s.}
    \label{figure:picca_recon_results}
\end{figure}

Rotation averaging was first proposed using the quaternion representation~\cite{DBLP:conf/cvpr/Govindu01}. 
Later solutions can be categorised into approaches based on either local and global optimisation. 
Local optimisation approaches such as the
one presented by Chatterjee and Govindu \cite{DBLP:conf/iccv/ChatterjeeG13} are well studied and practical. 
However, these methods only return the nearest local minimum. To overcome this limitation, the community has 
also proposed global optimisation approaches \cite{DBLP:conf/cvpr/TronZD16, rosen2019se, DBLP:conf/cvpr/ErikssonOKC18, 
EriksonOKC20,DBLP:conf/eccv/DellaertRWMC20}. Though the retrieval of global optima can be guaranteed, they have large 
computational cost and high sensitivity against outliers, and thus are impractical when applied to large-scale SfM problems.

In this paper, we focus on improving the efficiency and robustness of the rotation averaging method,
and on pushing its application to challenging scenes. We make a combination of a global solver and a 
local solver to solve rotation averaging, which guarantees global optimality and has strong resilience against outliers.
Rotation averaging based on chordal distances can be reformulated as a semi-definite program (SDP) 
with a low-rank constraint. 
Taking advantage of the low-rank factorisation of the original SDP, we can apply globally optimal Riemannian-Staircase-based
~\cite{DBLP:journals/corr/Boumal15} methods. In principle, any global solver~\cite{rosen2019se,
DBLP:journals/corr/abs-1903-00597, DBLP:conf/eccv/DellaertRWMC20} can be used in our hybrid approach, and 
we adopt the block coordinate minimisation method~\cite{DBLP:journals/corr/abs-1903-00597} to better leverage 
the view graph sparsity. By preprocessing the graph with fast view graph filtering, graph sparsity can be further exploited to accelerate the optimisation.

Previous works mainly apply rotation averaging to global SfM.
Though global SfM is efficient, translation averaging is often complicated by the unknown scale of relative translations and the difficulty of identifying outliers.
In this work, we embed the proposed approach into an incremental SfM pipeline.
The strategy is inspired by the work of Cui \emph{et al.}~\cite{DBLP:conf/cvpr/CuiGSH17}, who proposes a 
hybrid SfM scheme where camera rotations are estimated globally, and camera centers are estimated incrementally by a perspective-2-point (P2P) method. 
Though the approach is efficient, some estimated absolute rotations are not correct, thus it prevents camera centers from proper registration and scene reconstructions from completeness. Inspired by Cui's 
approach~\cite{DBLP:conf/cvpr/CuiGSH17}, we apply rotation averaging in a traditional incremental SfM pipeline,
however, use the perspective-3-point (P3P) to register camera rotations and centers. We further propose a novel cost function to optimise camera poses and landmarks that alleviates drift accumulation, in which camera rotations obtained from rotation averaging are used as regularisers. The resulting SfM approach, 
named RA-SfM (Rotation-Averaged Structure from Motion), shows high practicality and surpasses state-of-the-art
methods in accuracy, which is demonstrated at the hand of extensive experiments on large-scale 
real-world datasets. The reconstruction result of the largest dataset is shown in Fig.~\ref{figure:picca_recon_results}.

In summary, the main contributions of our work are:

\begin{itemize}
  \item We propose an outlier-resilient hybrid rotation averaging approach, which combines a global optimiser with fast 
  view graph filtering and a local optimiser.
  
  \item We refine the traditional incremental bundle adjustment cost function by adding the obtained global rotations as a regularisation term,
    which significantly alleviates drift accumulation in incremental SfM.
\end{itemize}

The practicality and superiority of the proposed scheme is demonstrated by extensive experiments on synthetic 
datasets and challenged internet datasets.

\section{Related Work}
\label{sec:related_work}

Motion averaging~\cite{DBLP:conf/cvpr/Govindu01,DBLP:conf/cvpr/Govindu04} is widely used in global 
SfM pipelines~\cite{DBLP:conf/iccv/MoulonMM13,DBLP:conf/iccv/CuiT15,DBLP:conf/iccv/SweeneySHTP15,
zhu2017parallel,DBLP:conf/cvpr/ZhuZZSFTQ18} as an answer to the drift problem occurring in incremental 
SfM~\cite{DBLP:conf/iccv/AgarwalSSSS09, DBLP:conf/3dim/Wu13,DBLP:conf/cvpr/SchonbergerF16,
DBLP:conf/3dim/CuiSGH17}. The first solution 
to rotation averaging goes back to Govindu \cite{DBLP:conf/cvpr/Govindu01}, who uses the quaternion 
representation and solves the problem by linear least-squares fitting. 
More reliable results were later on gained by optimising over a Lie algebra~\cite{DBLP:conf/cvpr/Govindu04}. 
In practice, the problem is complicated by the existence of outliers. To enhance the 
robustness of rotation averaging, absolute rotations may first be initialised under the $L_1$-norm, 
and then refined by Iteratively Reweighted Least Squares (IRLS) \cite{DBLP:conf/iccv/ChatterjeeG13,
DBLP:journals/pami/ChatterjeeG18}. Despite great progress, all aforementioned approaches can only guarantee a locally optimal solution. Another local approach was proposed by Crandall \emph{et al.} 
\cite{DBLP:conf/cvpr/CrandallOSH11,DBLP:journals/pami/CrandallOSH13}, who couple the cost function with 
regularisation terms to enhance robustness. However, the method is computationally demanding as it relies on discrete belief propagation over a Markov random field.

Fredriksson and Olsson \cite{DBLP:conf/accv/FredrikssonO12} exploit Lagrangian duality to become the 
first to find a globally optimal solution to the rotation averaging problem.
In a similar approach, Eriksson \emph{et al.} \cite{EriksonOKC20} perform the optimisation directly on 
the rotation matrix by minimising chordal distances. 
By removing the determinant constraint on the rotation from the original SDP, they 
elegantly prove that there is no duality gap between the primal problem and its dual when residual 
errors are bounded below an angular residual threshold. 

Rotation averaging can be converted into an SDP optimisation problem~\cite{DBLP:books/cu/BV2014}.
Wang and Singer \cite{DBLP:journals/ini/WangS13} solve it by 
the Alternating Direction Method of Multipliers (ADMM)~\cite{DBLP:journals/ftml/BoydPCPE11,
DBLP:journals/mpc/WenGY10}. Eriksson \emph{et al.} \cite{EriksonOKC20} use a row-by-row block 
coordinate descent method (BCM) \cite{Wen09rowby}. However, due to the slow convergence of ADMM and the repetitive fill-in procedures of BCM, neither approach proves to be practical when applied to large-scale datasets. A seminal work on the solution of SDP problems is presented by Burer and 
Monteiro~\cite{DBLP:journals/mp/BurerM03}, where the positive semi-definite variable is replaced by an appropriate factorisation, and the minimal rank variable is chosen to enhance computational speed. 
The Burer-Monteiro factorisation later inspired Boumal~\cite{DBLP:journals/corr/Boumal15}, who proposes 
a general optimisation technique named the Riemannian staircase algorithm, where the rank variable is 
augmented until the KKT condition is met, thus guaranteeing global optimality. 
Rosen \emph{et al.} ~\cite{rosen2019se} address the SDP problem of pose graph optimisation 
in the Special Euclidean space (SE(n)). When translation variables are decoupled from rotations, 
they first find the second-order critical point by the second-order Riemannian trust-region method,
and then adopt the low-rank optimisation framework of ~\cite{DBLP:journals/corr/Boumal15} to guarantee 
global optimality~\cite{rosen2019se}. Inspired by Wang's work~\emph{et al.}~\cite{DBLP:journals/corr/WangCK17},
which solves the low-rank SDP problem by block coordinate descent method,
Tian \emph{et al.}~\cite{DBLP:journals/corr/abs-1903-00597} extends this work to Steifel manifold,
and further applied a Riemannian BCM method to pose graph optimisation in distributed settings
~\cite{DBLP:journals/corr/abs-1911-03721}.
Building on SE-Sync~\cite{rosen2019se}, Dellaert \emph{et al.}~\cite{DBLP:conf/eccv/DellaertRWMC20}
propose \textit{Shonan rotation averaging}, a method in which the rotation matrix is vectorized, thus permitting 
the use of existing gradient-based optimisation methods 
on the manifold of rotation matrices.

\section{Notations and Preliminaries}
\label{sec:notation}

Let $\mathcal{G} = \{V, E\}$ be an undirected graph, where $V$ represents the collection of nodes and 
$E$ the set of edges. Let $m = |E|$ be the number of edges and $n = |V|$ be the number of nodes. 
Let $\tr(\cdot)$ denote the trace of a square matrix.
Given two matrices $A \in \mathbb{R}^{m\times n}$ and $B \in \mathbb{R}^{m\times n}$, let
 $\langle A, B \rangle = \sum_{i} \sum_j A_{ij}B_{ij}$. We therefore have 
 $\tr(A^TB) = \langle A, B \rangle$. 
Let $\blockdiag(A)$ represent the block diagonal matrix of $A$, and 
$\symblockdiag(A)=\frac{1}{2} \blockdiag(A + A^T)$.

The set of rotations in 3D forms the Special Orthogonal Group $\text{SO}(3)$, i.e.,
\begin{equation}
    \text{SO}(3) = \{R \in \mathbb{R}^{3 \times 3} | R^TR = I, \det(R) = 1\}.
\end{equation}
Since $\text{SO}(3)$ is a Lie group, there exists an exponential mapping between a rotation $R$ and 
its Lie algebra $\mathfrak{so}(3)$ representation $\bm{w}$~\cite{ma2012invitation}:
\begin{equation}
    R = \exp ([\bm{w}]_{\times}).
\end{equation}
The absolute rotations are grouped in $\mathcal{R} = \{R_1, R_2, \cdots, R_n\}$, 
where $R_i \in \text{SO}(3)$, $i \in [n]$. Relative rotations are represented 
by $\mathcal{R}_{\text{rel}} = \{R_{ij}\}$, where $R_{ij} \in \text{SO}(3)$, $i,j \in [n]$, $i < j$ 
is the rotation from $R_i$ to $R_j$. The chordal distance between two rotations is measured 
by~\cite{DBLP:journals/ijcv/HartleyTDL13}
\begin{equation}
\label{equ:chordal_d}
d_{\text{chord}}(R_1, R_2) = \left\|R_1 - R_2\right\|_F,
\end{equation}
where $\left\|\cdot\right\|_F$ represents the Frobenius norm of a matrix. 

\section{Hybrid Rotation Averaging}
\label{sec:RA_by_BM}

Globally optimal rotation averaging is sensitive to outliers, thus requiring an additional step to clean the view graph. In this section, we first present an efficient pre-processing step to filter outliers in the view graph. 
We then apply a block coordinate descent (BCD) method~\cite{DBLP:journals/corr/abs-1903-00597} to optimise the low-rank formulation of 
rotation averaging. Its global optimality can be guaranteed theoretically. Finally, we apply a local optimisation step to further refine the result in the case of scenes that have many erroneous edges.

\subsection{Fast View Graph Filtering}
\label{subsec:fast_view_graph_filtering}

The view graph plays an important role in our SfM pipeline. We clean the view graph for two main reasons:
(1) Solutions of global rotation averaging algorithms can be biased by outliers. Also, global optimality is only
guaranteed when the residuals for each edge are bounded below a certain threshold~\cite{EriksonOKC20}. (2) Some view pairs are redundant
and even harm the quality of SfM results.
Zach \emph{et al.} \cite{DBLP:conf/cvpr/ZachKP10} proposed a view graph filtering (VGF) technique to obtain a high-quality initial view graph, where loop constraints of rotation triplets are utilised to detect outliers. Specifically, edge $(i,j)$ is an outlier if its angular error 
below a given threshold $\epsilon$
\begin{equation}
\label{equ:loop_constraint}
    d(R_{ij}R_{jk}R_{ki}, I) > \epsilon.
\end{equation}
Despite its effectiveness, \cite{DBLP:conf/cvpr/ZachKP10} needs to validate all triplets, which is 
impractical for large-scale datasets.
However, \cite{DBLP:conf/eccv/ShenZFZQ16} suggests that it is not necessary to check all triplets to distinguish inliers from outliers, and that an increased number of valid 2D-2D image correspondences usually suggests more reliable two-view geometries. We propose an efficient view graph filtering method that relies on this observation. In the following, we denote a group of 3 nodes as a 
\emph{triplet}, and a triplet with two valid edges and one unverified edge as a \emph{weak triplet}.

Given an initial view graph $\mathcal{G}$, we start by constructing a maximum spanning tree (MST), 
where the weight of an edge is the number of valid 2D-2D correspondences. The relative rotations from this MST are all treated as valid. We then check the triplets along with the MST. That is, all adjacent edges that share a common node in the MST are used to build triplets. Next, we generate many weak triplets. Now supposing that edges $(i, j)$ and $(j, k)$ are valid and edge $(i, k)$ exists, we use 
criterion~\eqref{equ:loop_constraint} to verify the validity of edge $(i, k)$. An iteration is completed once all such weak triplets have been verified. After the first iteration, new weak triplets are generated based on which we can perform another iteration. We empirically found that $3$ 
iterations are sufficient for successful rotation averaging.

\subsection{Global Rotation Averaging}
\label{subsec:global_ra_optimization}

In this section, we first review a globally optimal guaranteed rotation averaging method
~\cite{DBLP:journals/corr/abs-1903-00597}, then
the sparsity pattern of the view graph is further exploited to accelerate the algorithm.

Given a set of relative rotations $\{R_{ij}\}$, where $i, j \in [n]$, the aim of rotation averaging is to obtain the absolute 
rotations $\{R_i\}$ that minimises the cost function below:
\begin{equation}
    \label{equ:rotation_averaging}
    \min_{R_1, \cdots, R_n}\ \sum_{(i,j) \in E} d^p(R_{ij}, R_j R_i^T),
\end{equation}
where $d^p(\cdot)$ represents a distance measure under a $p$-norm.
While there are a lot of local methods 
\cite{DBLP:conf/cvpr/Govindu01,DBLP:conf/cvpr/Govindu04,DBLP:conf/iccv/ChatterjeeG13,
DBLP:journals/pami/ChatterjeeG18} giving a least-squares solution to 
problem~\eqref{equ:rotation_averaging}, here we exploit a global optimisation approach that can obtain 
the global optimum.
Adopting the chordal distances, the primal problem of rotation averaging is finally given by
\footnote{See supplementary material for the complete derivation.}
\begin{equation}
\label{equ:primal_problem_ra}
\begin{split}
    &\min_R \ \ -\tr(R^TGR)
    \ \ \ \text{s.t.} \quad  R \in \text{SO}(3)^n,
\end{split}
\end{equation}
where $R = [R_1\ R_2\ \cdots\ R_n]$, and $G_{ij} = a_{ij} R_{ij}$, with $a_{ij} = 1$ if the edge 
between views $i$ and $j$ exists, and $0$ otherwise.
Eriksson \emph{et al.} \cite{EriksonOKC20} solve the dual problem with determinant constraint relaxation
to the primal problem~\eqref{equ:primal_problem_ra}
\begin{equation}
\label{equ:duals_dual_ra}
    \begin{split}
        \min_{X} \ \ &-\tr(GX) \\
        \text{s.t.} \quad &X_{ii} = I_3,\ i = 1, \cdots, n, \quad X \succeq 0,
    \end{split}
\end{equation}
where $X$ can be written as a block matrix $X_{ij}$, with $i, j \in [n]$ and 
$X_{ij} \in \mathbb{R}^3$.

Problem~\eqref{equ:duals_dual_ra} is an SDP problem \cite{DBLP:books/cu/BV2014}.
Since every $X \succeq 0$ can be factored as $Y^TY$ for some 
$Y$ \cite{DBLP:journals/mp/BurerM03}, and---in the case of rotation averaging---the optimal value of 
$X$ satisfies $X^{\star} = {R^{\star}}^T R^{\star}$~\cite{EriksonOKC20}, there is an implicit constraint on $X$ such that 
$\rank (X)=3$. Thus, $X$ can be reformulated as
\begin{equation}
\label{equ:low_rank_factorization}
    X = Y^T Y,
\end{equation}
where $Y = [Y_1\ Y_2\ \cdots\ Y_n], Y_i^T Y_i = I, \forall i \in [n]$. By substituting 
\eqref{equ:low_rank_factorization} into problem~\eqref{equ:duals_dual_ra}, a new problem is obtained
\begin{equation}
\label{equ:low_rank_sdp}
    \begin{split}
        \min_Y \ \ &-\tr(G Y^T Y) \qquad\qquad\\
        \text{s.t.} \quad &Y = [Y_1\ Y_2\ \cdots\ Y_n],\quad Y_i^T Y_i = I, \forall i \in [n].
    \end{split}
\end{equation}

While Tian \emph{et al.}~\cite{DBLP:journals/corr/abs-1903-00597} proposed a 
block coordinate minimisation (BCM) method to solve the low-rank SDP problem~\eqref{equ:low_rank_sdp}, 
we make a detailed derivation to the BCM solution, which is further accelerated by exploring graph sparsity. 
Note that
\begin{align}
\label{equ:trace_equation}
  &\tr(GY^TY)
  = \tr(YGY^T) = \langle YG, Y \rangle \\ \nonumber
  &= \sum_{j=1}^n \langle \sum_{i=1}^n Y_iG_{ij}, Y_j \rangle 
  = \sum_{j=1}^n \langle \hat{Q}_j, Y_j \rangle,
\end{align}
where $\hat{Q}_j = \sum_{i=1}^n Y_iG_{ij}$. Let $f(Y^k) = \sum_{j=1}^n \langle \hat{Q}_j^k, Y_j^k \rangle$, 
where superscript $k$ represents the $k$-th iteration in BCM. 
Since $G_{ii}=\mathbf{0}$, using \eqref{equ:trace_equation} we have
\begin{align}
  &\mathop{\arg\min}_Y f(Y^k) 
  = \mathop{\arg\min}_Y \sum_{j=1}^n \langle \hat{Q}_j^k, Y_j^k \rangle \nonumber \\
  = & \mathop{\arg\min}_Y \sum_{j=1}^n \langle \sum_{i \neq j}^n Y_i^k G_{ij}, Y_j^k \rangle 
  = \mathop{\arg\min}_Y \sum_{j=1}^n \langle Q_j^k, Y_j^k \rangle, \nonumber
\end{align}
where $Q_j = \sum_{i \neq j}^n Y_i G_{ij}$. This leads us to the derivation
\begin{align}
\label{equ:derivation_low_rank_bcm}
  & Y_{j_k}^{k+1}
  = \mathop{\arg\min}_{Y_{j_k}}\ f(Y_1^k, \cdots, Y_{j_k-1}^k, Y_{j_k}^k, Y_{j_k+1}^k, \cdots, Y_n^k) \nonumber\\
  = & \mathop{\arg\min}_{Y_{j_k}}\ \sum_{j=1}^n \langle \sum_{i \neq j}^n Y_i^k G_{ij}, Y_j^k \rangle \nonumber\\
  = & \mathop{\arg\min}_{Y_{j_k}}\ \langle Q_j^k, Y_{j_k}^k \rangle + \sum_{j \neq j_k}^n \sum_{i \neq j}^n \langle Y_i^k G _{ij}, Y_j^k \rangle \nonumber\\
  = & \mathop{\arg\min}_{Y_{j_k}}\ 2\langle Q_j^k, Y_{j_k}^k \rangle +  \sum_{j \neq j_k}^n \sum_{i \neq j,j_k}^n \langle Y_i^k G _{ij}, Y_j^k \rangle \nonumber\\
  = & \mathop{\arg\min}_{Y_{j_k}}\ 2\langle Q_j^k, Y_{j_k}^k \rangle 
  = \mathop{\arg\min}_{Y_{j_k}}\ \frac{1}{2} \left\|Y_{j_k}^k + Q_j^k\right\|_F^2.
\end{align}

By solving problem~\eqref{equ:derivation_low_rank_bcm}, the update of $Y_j$ in problem~\eqref{equ:low_rank_sdp} 
can be determined by~\cite{DBLP:journals/jscic/LaiO14, DBLP:journals/corr/abs-1903-00597}
\begin{equation}
\label{equ:svd_rotation}
    Y_j^{*} = U_j I_{3 \times 3} V_j^T = U_j V_j^T,
\end{equation}
where $U_j\Sigma V_j$ is the singular value 
decomposition of $-Q_j$.

Once the optimal value $Y_j^*$ is obtained, we need to update $Q_j$ at each inner iteration. 
The update rule is
\begin{align}
\label{equ:q_update_step}
  &  Q_j^{k+1}
  = \sum_{i \neq j}^n Y_i^{k+1}G_{ij}
  = Y_{j_k}^{k+1} G_{j_k j} + \sum_{i \neq j, j_k}Y_i^{k+1}G_{ij} \nonumber\\
  = & Y_{j_k}^{k+1} G_{j_k j} + \sum_{i \neq j, j_k} Y_i^k G_{ij} + Y_{j_k}^k G_{j_k j} - Y_{j_k}^k G_{j_k j} \nonumber\\
  = & Y_{j_k}^{k+1} G_{j_k j} + \sum_{i \neq j} Y_i^k G_{ij} - Y_{j_k}^k G_{j_k j} \nonumber\\
  = & Q_j^k + (Y_{j_k}^{k+1} - Y_{j_k}^k) G_{j_k j}.
\end{align}

In Algorithm~\ref{alg:rotation_averaging_algorithm}, we outline the BCM with graph sparsity. 
In steps 6$\sim$7 of Algorithm~\ref{alg:rotation_averaging_algorithm}, the time complexity of $O(n)$ is only an upper bound occurring for general cases. In practice, due to the commonly sparse structure of SfM problems, \textit{time complexity can be further reduced to $O(d)$}, where $d$ is the degree of the nodes. This property is important for accelerating the optimisation. Notice that,
with our fast view graph filtering, the graph can be more sparse and we can gain more acceleration.

\begin{algorithm}
  \caption{BCM for SDP~\cite{DBLP:journals/corr/abs-1903-00597} with Graph Sparsity}
  \label{alg:rotation_averaging_algorithm}
  \begin{algorithmic}[1]
    \Require relative rotations $\mathcal{R}_{\text{rel}}$, $maxIterNum$, $Y^0$.
    \Ensure First-order critical point $Y^{\star}$
    
    \State $k \leftarrow 0$; 
    $Q_j^0 \leftarrow \sum_{i \neq j}^n Y_i G_{ij}, \forall j \in [n]$.
    
    \While{$k < maxIterNum$ AND not converge}
    \For{$i < n$}
    \State $j_k \leftarrow i$
    \State Update $Y_{j_k}^{k+1}$ by Eq.~\eqref{equ:svd_rotation}
    \For{$\forall j \neq j_k$ AND $G_{j_k j} \neq \mathbf{0}$  }
    \State Update $Q_j^{k+1}$ by Eq.~\eqref{equ:q_update_step}
    \EndFor
    \EndFor
    
    \State $k \leftarrow k + 1$;
    \EndWhile
    
    \State return $Y$
    
  \end{algorithmic}
\end{algorithm}

\paragraph{Discussion of Global Optimality: } Problem~(\ref{equ:low_rank_sdp}) is non-convex, 
and there is no guarantee that we can obtain the global optimum. 
In~\cite{DBLP:journals/corr/WangCK17}, the global optimum is guaranteed by selecting an appropriate
step size and random initialisation (Theorem 3.4). However, the theorem only holds for scalar variables. For the RA problem, global optimality is not held since we optimise on the manifold.
Boumal~\cite{DBLP:journals/corr/Boumal15} proposes a general framework named the 
\textit{Riemannian Staircase} algorithm (RS), which can find the global optimum. 
As previous work has applied a Riemannian based method to ensure the global optimum
~\cite{rosen2019se, DBLP:journals/corr/abs-1911-03721, DBLP:conf/eccv/DellaertRWMC20}, 
we recommend interested readers to refer to them.

\subsection{Local Optimization Refinement}
\label{subsec:local_refine}

The global optimisation presented in Sec.~\ref{subsec:global_ra_optimization} assumes that the input relative rotations do not contain any outliers. As a result, it is sensitive to outliers.
To further improve the robustness and accuracy of the low-rank BCM method,
we follow the \emph{suggest-and-improve} framework of~\cite{park2017general}.
The global optimisation approaches can obtain a good solution close to the global optimum. Still, it could be further refined by a gradient descent algorithm.
We adopt this framework and use the method of Chatterjee and Govindu \cite{DBLP:conf/iccv/ChatterjeeG13} 
as a local optimiser. This method performs an Iteratively Reweighted Least Square (IRLS) under Lie algebra, 
which leads to an efficient and robust optimiser. The rotation $R_{ij}, R_i, R_j$ can be represented by the 
corresponding Lie algebra $\boldsymbol{\omega}_{ij}, \boldsymbol{\omega}_i, \boldsymbol{\omega}_j$, respectively. 
Using the Baker-Campbell-Hausdorff (BCH) equation, a single constraint in Eq.~\eqref{equ:rotation_averaging} can 
be converted to
\begin{equation}
\label{equ:lie_algebra_ra_constraint}
    \boldsymbol{\omega}_{ij} = \boldsymbol{\omega}_j - \boldsymbol{\omega}_i.
\end{equation}
By collecting the relative constraints, we obtain
\begin{equation}
\label{equ:linear_ra_equation}
    A \boldsymbol{\omega}_{\text{global}} = \boldsymbol{\omega}_{\text{rel}}.
\end{equation}
Here $A$ is a sparse matrix, in which all consecutive $3 \times 3$ blocks are zeros except two matrices $I$ and $-I$. 
Encapsulating Eq.~\eqref{equ:linear_ra_equation} with a Huber loss $\rho(x) = \frac{x^2}{x^2 + \sigma^2}$, 
we can optimise a robust cost function under least square meaning

\begin{equation}
\label{equ:robust_l2}
    \mathop{\arg \min}_{\boldsymbol{w}_{\text{global}}} \sum \rho(\left\|A \boldsymbol{\omega}_{\text{global}} 
    - \boldsymbol{\omega}_{\text{rel}}\right\|).
\end{equation}

\subsection{Hybrid of Global Rotation Averaging}

\begin{figure}[htbp]
  \centering
  \includegraphics[width=0.95\linewidth]{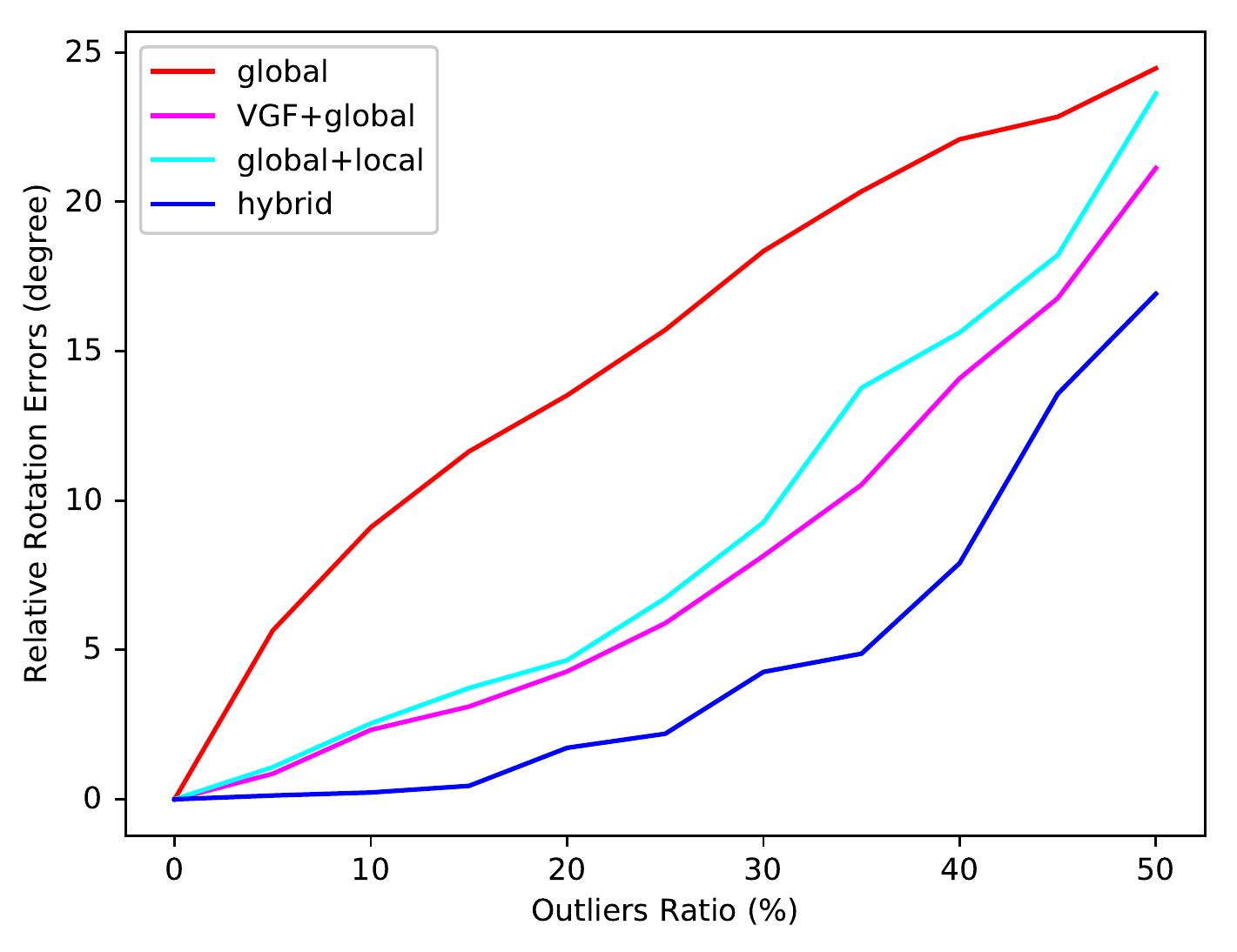}
  \caption{Ablation study of robustness. Outliers are generated by perturbing ground truth by 
    rotation between $60^{\circ}-90^{\circ}$. \textit{global} represents the proposed low-rank BCM 
    method, and \textit{hybrid} represents the proposed hybrid method with VGF and local refinement.}
  \label{fig:outlier_robustness}
\end{figure}

We outline our hybrid rotation averaging algorithm in Algorithm ~\ref{alg:hybrid_rotation_averaging_algorithm}.
An ablation study of robustness against outliers is shown in Fig.~\ref{fig:outlier_robustness}. The outlier ratio ranges from $0$ to $50\%$ 
and is incremented in steps of $5\%$. We display the mean rotation error over 30 experiments. 
As can be observed, both VGF and local refinement improve the robustness of the global rotation averaging approach.

\begin{algorithm}[htbp]
  \caption{Hybrid Rotation Averaging Algorithm}
  \label{alg:hybrid_rotation_averaging_algorithm}
  \begin{algorithmic}[1]
  \Require relative rotations $\mathcal{R}_{\text{rel}}$
  \Ensure global rotations $\mathcal{R} = \{R_1, R_2, \cdots, R_n\}$
  
  \State Perform fast VGF as described in Sec.~\ref{subsec:fast_view_graph_filtering}.
  
  \State Calculate global rotations using Algorithm~\ref{alg:rotation_averaging_algorithm} (or any other global rotation averaging method).
  
  \State Refine global rotations by solving problem~\eqref{equ:robust_l2}.
  
  \end{algorithmic}
\end{algorithm}

\section{Rotation-Averaged Structure from Motion}
\label{sec:ra_application}

In this section, we apply our rotation averaging method to an incremental SfM pipeline, which is known to suffer from the drift problem. Our hybrid SfM pipeline can be summarised as follows:
We first construct the view graph and obtain global rotations from our proposed hybrid rotation averaging approach. Next, we create a seed reconstruction by selecting two appropriate images. We then continue by 
incrementally registering adjacent camera poses using a
RANSAC-based~\cite{DBLP:journals/cacm/FischlerB81} P3P~\cite{DBLP:conf/cvpr/KneipSS11} 
algorithm, and triangulate landmarks. To reduce the accumulation of errors in our incremental SfM pipeline, 
we perform local bundle adjustment after each successful registration of an image, and global bundle adjustment whenever the number of recently added views surpasses a certain threshold.

The drift problem is not solved, as each newly computed camera pose is affected by a small error, and these errors accumulate along the graph. Traditional incremental SfM pipelines have no way to rectify these errors. 
To tackle this problem, we introduce a novel cost function with averaged rotations as regularisers for bundle adjustment. 
Let $\mathcal{I}_i$ denote the measurements of image $i$. 3D landmarks observed by image $i$ are denoted as set $\mathcal{P}_i$. 
Note that the sets $\{\mathcal{P}_{i} | i \in \mathcal{I}\}$ might have repetitive elements, 
which---in a slight abuse of notation---is ignored for the sake of simplicity.
Let $\mathbf{u}_{il}\in\mathcal{I}_i$ furthermore denote the image keypoint measurement of 
landmark $l$ in frame $i$. the pre-computed known rotation with respect to image $i$ is denoted 
as $\hat{\mathcal{R}}_{i}$.

The proposed cost function is given by
\begin{align}
\label{eq:map_joint_opt}
  \sum_{i \in \mathcal{I}} \sum_{l \in \mathcal{I}_i} \rho_v \left( 
  \left\| \mathbf{r}_{\mathcal{I}_{il}} \right\|^2 \right) 
+ \sum_{(i,j) \in \mathcal{E}} w_{ij} \left(
  \left\| \mathbf{r}_{\mathcal{R}_{ij}} \right\|_{}^2 \right),
\end{align}
where $\rho_v (\cdot)$ is a robust loss function, and $w_{ij}$ is an individual weighting function for 
each known rotation term. In this paper, we fix $w_{ij}$ as a constant. 
The objective divides into two terms which are explained as follows.

{\bf Visual Term:}
We adopt the traditional re-projection error in bundle adjustment as our visual term
\begin{equation}
\mathbf{r}_{\mathcal{I}_{il}} = \mathbf{u}_{il} - \mathbf{\Pi} (R_i, C_i, \mathcal{P}_i, l),
\end{equation}
where $R_i$ and $C_i$ are respectively the estimated camera rotation and center, and $ \mathbf{\Pi}(\cdot)$ 
is the back-projection function that projects landmarks into the image plane. 
Note that the latter also depends on camera intrinsics, which---for the sake of a simplified and general 
notation---are not specified.

{\bf Known Rotation Term:}
The added known rotation term is
\begin{equation}
\mathbf{r}_{\mathcal{R}_{ij}} = \log(
 \hat{\mathcal{R}}_j^T \hat{\mathcal{R}}_{i}  R_i^T R_j),
\end{equation}
where $\log$ is the logarithm map $\mathrm{SO}(3) \rightarrow \mathfrak{so}(3)$. 
This known rotation term is used as a regulariser in the complete cost function.

\begin{figure}[!ht]
  \centering
  \subfigure[Incremental camera registerisation.]
  {
     \begin{minipage}{0.8\linewidth}
         \includegraphics[width=1\linewidth]{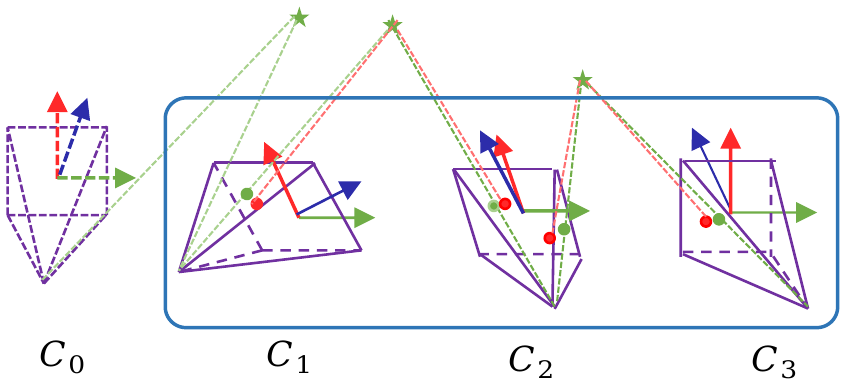}
     \end{minipage}
  }

  \subfigure[Incremental camera registration with averaged rotations.]
  {
      \begin{minipage}{0.82\linewidth}
          \includegraphics[width=1\linewidth]{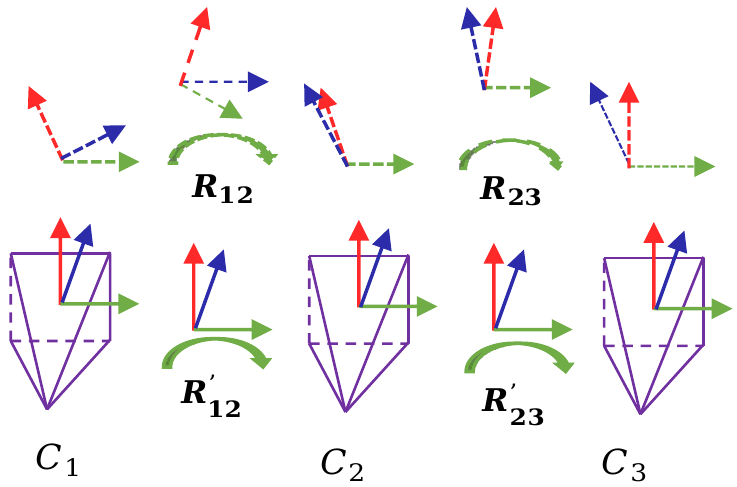}
      \end{minipage}
  }
      
  \caption{A toy example to explain our RA-SfM. Camera rotations are drawn by three arrows with colors 
  in red, green, and blue. In (a), landmarks are denoted by $\star$, green dots in the image plane represent 
  keypoints, and red dots are reprojected coordinates. In (b), we show the correctly registered camera poses for
  $C_1, C_2, C_3$. $R_{12}$ and $R_{23}$ denote the relative rotations, which are obtained from (a). 
  $R_{12}^{'}$ and $R_{23}^{'}$ denote the relative rotations obtained from averaged rotations.}
  \label{figure:incremental_ba_ra}
\end{figure}

To better demonstrate the effectiveness of~\eqref{eq:map_joint_opt}, we further make an explanation for 
our cost function, and draw it as a toy example in Fig.\ref{figure:incremental_ba_ra}. In Eq.
~\eqref{eq:map_joint_opt}, the first term corresponds to the reprojection error, 
the second term is used to penalize the large RPE that is caused by pose drift. Note that the reprojection error 
may remain small even if camera poses are drifting. This can be seen from Fig.\ref{figure:incremental_ba_ra} (a): 
Suppose $C_0$ is correctly registered at first, and $C_1, C_2, C_3$ are registered sequentially. When $C_1$ is
wrongly registered, the error will be passed to $C_2$, then $C_3$. But traditional BA can not optimise camera poses
to the correct place, because the triangulated 3D points are geometrically coherent with the camera poses, and 
reprojection errors are small (measured by the red and green dots in Fig.\ref{figure:incremental_ba_ra}). 
However, in this situation, the known rotation term can measure
the discrepancy between the averaged rotations and incrementally recovered rotations
(See from Fig.\ref{figure:incremental_ba_ra} (b)). And our
optimiser tries to minimise this discrepancy and thereby alleviate pose drift.

\section{Experimental Results}
\label{sec:experiment}
Our experiments aim at demonstrating the accuracy, efficiency, and robustness of the proposed methods.
We implement Levenberg-Marquardt (LM)~\cite{DBLP:books/sp/NocedalW99},
row-by-row block coordinate descent (RBR-BCD)~\cite{EriksonOKC20}, 
and our hybrid rotation averaging in C++. Besides, the implementations of SE-Sync~\cite{rosen2019se} and 
Shonan~\cite{DBLP:conf/eccv/DellaertRWMC20} are provided by the authors and publicly available.
For HSfM~\cite{DBLP:conf/cvpr/CuiGSH17} and LUD~\cite{DBLP:conf/cvpr/OzyesilS15}, we use
\cite{DBLP:journals/pami/ChatterjeeG18} as the rotation averaging solver,
and the Ceres solver~\cite{ceres-solver} for bundle adjustment. 
All approaches are tested on a laptop with a 2.7 GHz CPU and 8GB RAM.

\begin{table*}[t]
  \centering
  \caption{Comparison of runtime on synthetic datasets. $n$ is the number of rotations, 
          and $\bar{R}$ represents the average rotation error (unit: degree).}
  \vspace{0.05in}
  \resizebox{0.95\textwidth}{!}{
  \begin{tabular}{| c || c || c || c | c || c | c || c | c || c | c || c | c |}
        \hline
        
        \multirow{2}{*}{$n$} & \multirow{2}{*}{\#edges} 
                             & \multirow{2}{*}{$\sigma$}
                             & \multicolumn{2}{c|}{\textbf{LM}~\cite{DBLP:books/sp/NocedalW99}}
                             & \multicolumn{2}{c|}{\textbf{RBR-BCD}~\cite{EriksonOKC20}}
                             & \multicolumn{2}{c|}{\textbf{Shonan}~\cite{DBLP:conf/eccv/DellaertRWMC20}}
                             & \multicolumn{2}{c|}{\textbf{SE-Sync}~\cite{rosen2019se}}
                             & \multicolumn{2}{c|}{\textbf{hybrid RA} } \\
        
        \cline{4-13} & \ & \ & $\bar{R}$ & $time(s)$
                            & $\bar{R}$ & $time(s)$
                            & $\bar{R}$ & $time(s)$
                            & $\bar{R}$ & $time(s)$
                            & $\bar{R}$ & $time(s)$ \\
        \hline
        
        \multirow{2}{*}{20} 
        & \multirow{2}{*}{30} & 0.2 
        & 5.201e-05 & 0.001
        & 1.219e-05 & 0.039
        & 9.803e-06 & 0.032
        & 8.975e-06 & $<$ \textbf{1e-06}
        & \textbf{7.307e-06} & $<$ \textbf{1e-06} \\
        \cline{3-13} & \ & 0.5 
        & 1.492e-01 & \textbf{0.002}
        & 7.200e-02 & 0.028
        & 1.955e-01 & 0.038
        & 2.033e-01 & 0.003
        & \textbf{1.550e-01} & \textbf{0.002}  \\
        \hline
        
        \multirow{2}{*}{100}
        & \multirow{2}{*}{300} & 0.2 
        & 9.490e-06 & 0.095
        & 6.351e-06 & 0.651 
        & \textbf{5.374e-06} & 0.144
        & 5.383e-06 & \textbf{0.005}
        & 5.938e-06 & 0.006\\
        \cline{3-13} & \ & 0.5 
        & 8.771e-01 & 0.089
        & 1.160e-01 & 0.813
        & 1.108e-01 & 0.078
        & 1.336e-01 & \textbf{0.079}
        & \textbf{9.400e-02} & 0.189 \\
        \hline
        
        \multirow{2}{*}{500} 
        & \multirow{2}{*}{1000} & 0.2 
        & 5.413e-06  & 0.845  
        & 5.381e-06 & 208.372
        & 5.433e-06 & 0.886
        & 5.209e-06 & 0.403
        & \textbf{5.184e-06} & \textbf{0.159} \\
        \cline{3-13} & \ & 0.5 
        & 7.351e-01  & 0.781  
        & 1.130e-01 & 245.818
        & 1.255e-01 & 0.533
        & \textbf{9.417e-02} & \textbf{0.255}
        & 1.080e-01 & 0.270 \\
        \hline
        
        \multirow{2}{*}{1000} 
        & \multirow{2}{*}{4000} & 0.2 
        & 1.800e-01 & 1.213 
        & \textbf{6.754e-06} & 2,274
        & 9.835e-06 & 1.883
        & 8.739e-06 & 0.821
        & 1.021e-05 & \textbf{0.127}\\
        \cline{3-13} & \ & 0.5 
        & 8.956e-01 & 1.372 
        & 1.120e-01 & 2,153 
        & 8.974e-02 & 1.141
        & 1.310e-01 & 0.985 
        & \textbf{8.200e-02} & \textbf{0.949} \\
        \hline
        
        \multirow{2}{*}{5000} 
        & \multirow{2}{*}{20000} & 0.2 
        & 1.260e-01 & 4.414 
        & - & -
        & 8.341e-06 & 14.870
        & \textbf{6.371e-06} & 9.083
        & 7.159e-06 & \textbf{0.331} \\
        \cline{3-13} & \ & 0.5 
        & 2.787e-01 & 5.183 
        & - & -
        & 1.516e-01 & 12.338
        & 1.408e-01 & 3.699
        & \textbf{8.200e-02} & \textbf{0.809} \\
        \hline
        
        \multirow{2}{*}{10000} 
        & \multirow{2}{*}{40000} & 0.2 
        & 1.410e-01 & 23.714 
        & - & - 
        & 1.838e-05 & 45.680
        & 9.037e-06 & 10.335 
        & \textbf{7.884e-06} & \textbf{0.362} \\
        \cline{3-13} & \ & 0.5 
        & 3.240e-01 & 27.265
        & -  & -
        & 1.209e-01 & 42.627
        & 1.386e-01 & 11.128 
        & \textbf{9.100e-02} & \textbf{1.704} \\
        \hline
        
        \multirow{2}{*}{50000} 
        & \multirow{2}{*}{200000} & 0.2 
        & - & - 
        & - & -
        & 6.013e-06 & 956.821 
        & - & -
        & \textbf{6.310e-06} & \textbf{0.515} \\
        \cline{3-13} & \ & 0.5 
        & - & - 
        & - & - 
        & 1.933e-01 & 905.294
        & - & -
        & \textbf{7.500e-02} & \textbf{7.124} \\
        \hline
  \end{tabular}
  }
  \label{table:synthetic_data}
\end{table*}

\subsection{Evaluation of Hybrid Rotation Averaging on Synthetic Datasets}
We designed 7 synthetic datasets to evaluate the performance of our rotation averaging approach, 
The view and relative rotation numbers are shown in Table~\ref{table:synthetic_data}, and denoted by
$n$ and $\#\text{edges}$ respectively.
The ground truth absolute rotations are initialised randomly. The relative rotations are constructed by a spanning tree expanded by random edges until the given number of relative poses is reached. All relative rotations are derived from ground truth, and perturbed by 
random angular rotations about randomly selected axes.
The perturbation angles are normally distributed with $0$ mean and variance of either $\sigma=0.2$
~rad or $\sigma = 0.5$~rad. Initial absolute rotations are chosen randomly.

The evaluation results are shown in Table~\ref{table:synthetic_data}, where we compare our method against 
LM~\cite{DBLP:books/sp/NocedalW99}, RBR-BCD~\cite{EriksonOKC20}, Shonan~\cite{DBLP:conf/eccv/DellaertRWMC20} and 
SE-Sync~\cite{rosen2019se}. In terms of efficiency, RBR-BCD is the slowest and almost 1000 times slower
than others when $n = 1000$. SE-Sync is faster than LM but stays within the same order of magnitude.
While SE-Sync is slightly faster than ours when the camera's number is below 500, the hybrid rotation averaging approach is $1\sim 2$ orders of magnitude faster than SE-Sync when the number of views
grows beyond 1000. While Shonan is also a low-rank method, as well as SE-Sync and ours, it is almost 
4 times slower than SE-Sync, and $2\sim3$ orders of magnitude slower than ours, when the number of the cameras goes above 5000. 
In terms of the scale of the solved problems, 
RBR-BCD failed when the camera number increased to $5000, 10000,$ or $50000$. This is primarily due to 
insufficient memory for optimisation, and we marked the corresponding cells in the table by ``--''. LM and 
SE-Sync failed when the camera number reaches $50000$, as there is insufficient memory to perform the CHOLMOD
~\cite{DBLP:journals/toms/ChenDHR08} factorisation.
As our approach only needs to compute the SVD of a small block matrix and evaluate $d$ matrix operations 
of $3 \times 3$ matrices in each iteration, we can solve all the evaluated large-scale datasets.

In terms of accuracy, LM achieves the global optimum with 
certain probability ($30\%-70\%$ as reported in \cite{EriksonOKC20}), and Table~\ref{table:synthetic_data} 
only shows the best results. While all the evaluated globally optimal methods have the same accuracy and can both obtain the global optimum for successful cases.

\subsection{Evaluation of RA-SfM on Real-World Datasets}

We evaluate the performance of our RA-SfM on large scale real-world datasets and compare it against 
state-of-the-art incremental~\cite{DBLP:conf/cvpr/SchonbergerF16}, global~\cite{DBLP:conf/cvpr/OzyesilS15} 
and hybrid~\cite{DBLP:conf/cvpr/CuiGSH17} SfM approaches. Since the quasi-convex SfM 
approach~\cite{DBLP:conf/cvpr/ZhangCL18} is sensitive to outlies and extremely slow in such datasets,
we did not evaluate it in our experiment.

\begin{figure}[t]
  \centering
  {
    \includegraphics[width=0.99\linewidth]{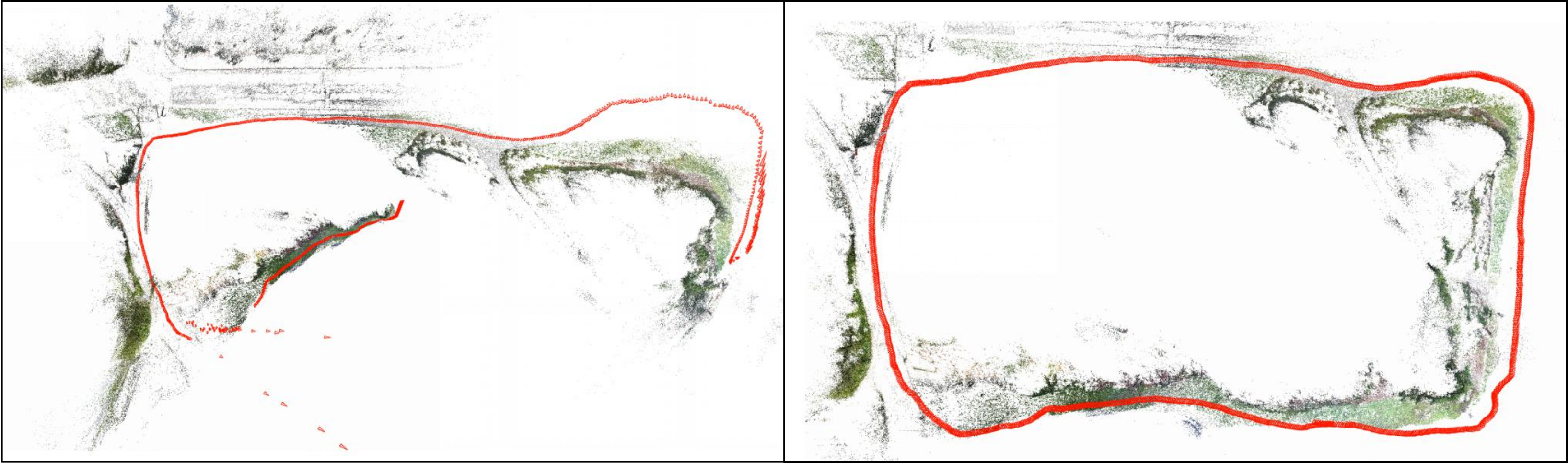}
  }    
  \caption{Reconstruction results for the Campus dataset~\cite{DBLP:conf/iccv/CuiT15}. 
           Left: COLMAP~\cite{DBLP:conf/cvpr/SchonbergerF16}, Right: Our RA-SfM.}
  \label{figure:campus_result}
  \vspace{-0.2in}
\end{figure}

Figure~\ref{figure:campus_result} shows the reconstruction results of COLMAP and our RA-SfM on the
Campus~\cite{DBLP:conf/iccv/CuiT15} dataset. This dataset, which has a loop, mainly contains plants that can produce lots of wrong matching results. COLMAP~\cite{DBLP:conf/cvpr/SchonbergerF16}
fails to reconstruct this dataset, as the camera poses drift and the loop is not closed. Our
approach closes the loop successfully, as the known rotation optimisation can further constrain
camera poses after the initial registration.

\begin{table*}[!ht]
  \centering
  \caption{Comparison of runtime and accuracy on online datasets \cite{DBLP:conf/eccv/WilsonS14}. 
    $N_i$ and $N_c$ denotes the number of images and registered cameras respectively, 
    $N_p$ denotes the reconstructed 3D landmarks, MRE means mean reprojection error in pixel. $T$ and $T_R$ 
    denotes the total reconstruction time and hybrid rotation averaging time respectively (in seconds).
    The best MREs are marked in bold font.}
  \vspace{0.05in}
  \resizebox{1.0\textwidth}{!}{
    \begin{tabular}{| c |
                    | c |
                    | c | c | c | c |
                    | c | c | c | c |
                    | c | c | c | c |
                    | c | c | c | c | c |}
      \hline
      
      \multirow{2}{*}{Dataset}  &
      \multirow{2}{*}{$N_i$} &
      \multicolumn{4}{c||}{\textbf{COLMAP} \cite{DBLP:conf/cvpr/SchonbergerF16}} & 
      \multicolumn{4}{c||}{\textbf{LUD} \cite{DBLP:conf/cvpr/OzyesilS15}} &
      \multicolumn{4}{c||}{\textbf{HSfM} \cite{DBLP:conf/cvpr/CuiGSH17}} & 
      \multicolumn{5}{c|}{\textbf{RA-SfM}}\\
      \cline{3-19} & \ & $N_c$ & $N_p$ & MRE & $T (s)$ 
                       & $N_c$ & $N_p$ & MRE & $T (s)$ 
                       & $N_c$ & $N_p$ & MRE & $T (s)$ 
                       & $N_c$ & $N_p$ & MRE & $T_R (s)$ & $T (s)$\\
      \hline
        Alamo & 2,915 
        & 906 & 138K  & 0.69 & 3,180            
        & 578 & 146K & 1.28 & 260               
        & 522 & 149K & 1.62 & 1,079             
        & 895 & 141K & \textbf{0.65} & 0.340 & 2,771 \\   
        \hline
        
        Ellis Island & 2,587 
        & 801 & 154K & 0.73 & 4,307             
        & 234 & 16K  & 1.54 & 24                
        & 208 & 34K  & 2.53 & 169               
        & 727 & 146K & \textbf{0.72} & 2.290 & 3,920 \\    
        \hline
        
        Gendarmenmarkt & 1,463 
        & 1,040 & 209K & 0.71 & 3,737            
        & 705   & 87K & 1.51 & 104              
        & 542   & 74K & 1.94 & 377              
        & 1,023 & 202K & \textbf{0.70} & 1.997 & 3,931 \\  
        \hline
        
        Madrid Metropolis & 1,344 
        & 460 & 60K & 0.62 & 1,320              
        & 350 & 51K & 1.08 & 36                 
        & 292 & 51K & 1.48 & 221                
        & 438 & 66K & \textbf{0.59} & 0.420 & 1,417 \\     
        \hline
        
        Montreal N.D. & 2,298 
        & 554 & 107K & \textbf{0.67} & 1,902             
        & 462 & 166K & 1.64 & 194               
        & 418 & 155K & 1.95 & 1041              
        & 528 & 105K & 0.68 & 0.115 & 1,423 \\    
        \hline
        
        Notre Dame & 1,431 
        & 1,408 & 349K & 0.76 & 22,788           
        & 550 & 262K & 2.06 & 259               
        & 526 & 281K & 2.30 & 2,375             
        & 1,409 & 353K & \textbf{0.75} & 0.131 & 19,943 \\  
        \hline
        
        NYC Library & 2,550 
        & 556 & 101K & 0.72 & 1,698             
        & 336 & 70K  & 1.52 & 75                
        & 282 & 74K  & 1.99 & 356               
        & 519 & 100K & \textbf{0.65} & 0.515 & 1,715 \\    
        \hline
        
        Piazza del Popolo & 2,251 
        & 1,011 & 122K & 0.68 & 2,676            
        & 329  & 38K  & 1.65 &  62              
        & 286  & 35K  & 1.92 & 212              
        & 966  & 122K & \textbf{0.66} & 0.360 & 3,258 \\   
        \hline
        
        Piccadilly & 7,351
        & 3,129 & 362K & \textbf{0.73} & 16,590            
        & 2,301 & 202K & 1.83 & 262               
        & 1,665 & 185K & 2.09 & 2,169             
        & 3,041 & 363K & 0.80 & 1.422 & 15,109 \\  
        \hline
        
        Roman Forum & 2,364 
        & 1,594 & 284K & \textbf{0.71} & 5,388             
        & 1,045 & 256K & 1.71 & 182               
        & 1,071 & 262K & 1.93 & 2,237             
        & 1,460 & 267K & 0.77 & 1.938 & 5,408 \\    
        \hline
        
        Tower of London & 1,576 
        & 707 & 140K & 0.61 & 2,767               
        & 485 & 140K & 1.65 & 95                  
        & 398 & 149K & 1.91 & 816                 
        & 672 & 139K & \textbf{0.58} & 0.800 & 1,979 \\      
        \hline
        
        Trafalgar & 15,685 
        & 6,980 & 581K & 0.81 & 14,790            
        & 5,044 & 378K & 1.56 & 713               
        & 3,446 & 318K & 1.95 & 5,761             
        & 7,085 & 597K & \textbf{0.72} & 4.213 & 14,831 \\  
        \hline
        
        Union Square & 5,961 
        & 937 & 69K & 0.66 & 2,604                
        & 803 & 41K & 1.65 & 107                  
        & 769 & 38K & 1.88 & 1,763                
        & 809 & 57K & \textbf{0.52} & 1.304 & 1,962 \\       
        \hline
        
        Vienna Cathedral & 6,288 
        & 1,185 & 290K & 0.74 & 9,714             
        & 849 & 203K & 1.91 & 173                 
        & 662 & 252K & 2.36 & 2,307               
        & 1,173 & 303K & \textbf{0.71} & 1.959 & 16,111 \\    
        \hline
        
        Yorkminster & 3,368 
        & 1,022 & 259K & 0.71 & 10,806            
        & 421 & 132K & 1.75 & 135                 
        & 417 & 129K & 1.93 & 1,487               
        & 614 & 183K & \textbf{0.64} & 3.183 & 9,299 \\        
        \hline
        
      \end{tabular}
    }
    \label{table:internet_data}
  \end{table*}

\begin{figure*}[t]
  \centering
  {
    \includegraphics[width=0.9\linewidth]{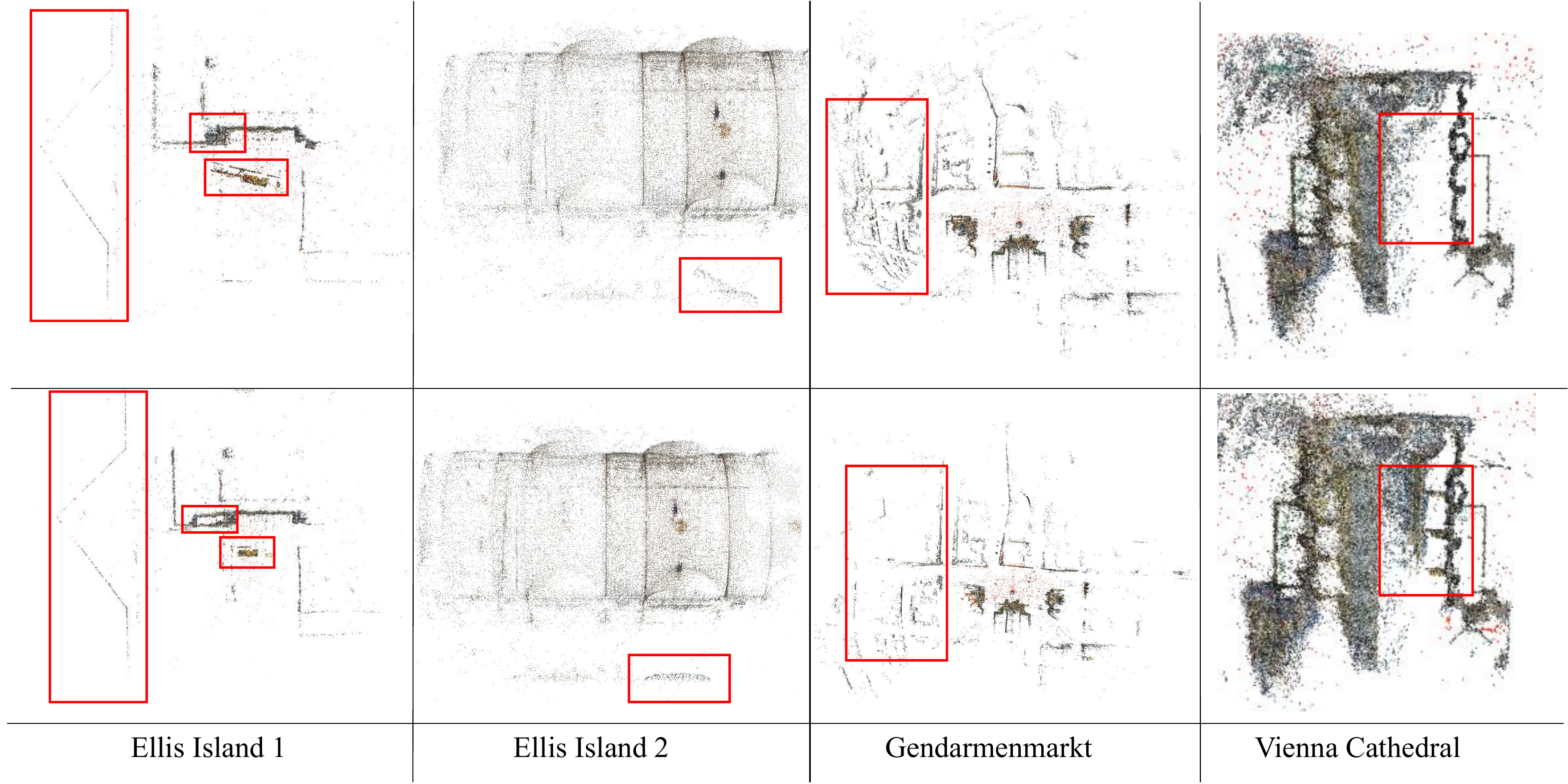}
  }
  \caption{Visual reconstruction results for some of the online datasets~\cite{DBLP:conf/eccv/WilsonS14}.
  For each subfigure, the top and bottom images are respectively the results obtained from 
  COLMAP~\cite{DBLP:conf/cvpr/SchonbergerF16} and our RA-SfM. (The first two columns are results of two parts of the Ellis Island dataset.)
  }
  \label{fig:visual_internet_recon_results}
  \vspace{-0.15in}
\end{figure*}

We also evaluated our approach on the online datasets from~\cite{DBLP:conf/eccv/WilsonS14}, which are collections of challenging unordered images. The datasets contain many wrong epipolar geometries 
due to the extreme viewpoint, scale, and illumination changes.
The runtime and accuracy results are shown in Table~\ref{table:internet_data}. As is observed,
COLMAP~\cite{DBLP:conf/cvpr/SchonbergerF16} recovers the most camera poses in most online datasets.
However, our method has the lowest mean reprojection error (MRE) in most of the online datasets,
which indicates RA-SfM is more robust and accurate than COLMAP.
For our RA-SfM, the time for rotation averaging is separately given in the penultimate column (denoted as
$T_R$). While LUD~\cite{DBLP:conf/cvpr/OzyesilS15}
is the most efficient one among the evaluated methods, it has large MRE, and the number of recovered
camera poses is less than ours and COLMAP. HSfM~\cite{DBLP:conf/cvpr/CuiGSH17} is faster
than COLMAP and RA-SfM because it only samples 2 correspondences to compute the camera centers in 
each RANSAC iteration. Besides, HSfM~\cite{DBLP:conf/cvpr/CuiGSH17} recovers the fewest camera poses and 
fails to recover the correct camera centers.

Some visual results for online datasets are shown in Fig.~\ref{fig:visual_internet_recon_results}.
For each subfigure, the top and bottom images are the results obtained by 
COLMAP~\cite{DBLP:conf/cvpr/SchonbergerF16} and our RA-SfM, respectively. 
For the Ellis Island dataset, we showed two different parts in the first two columns of 
Fig.~\ref{fig:visual_internet_recon_results}, where the red rectangle area shows the comparison result.
For the Gendarmenmarkt dataset, the reconstruction result of COLMAP is bad on the left part, which
indicates the wrong camera poses. For the Vienna Cathedral dataset, though COLMAP recovers more camera poses than
ours, our approach reconstructed more scene details than COLMAP, as is indicated by the red rectangle.
From Table~\ref{table:internet_data} and Fig.~\ref{fig:visual_internet_recon_results}, we demonstrate  
our RA-SfM can effectively correct the wrongly registered camera poses in the incremental SfM pipeline, 
and also achieves the state-of-the-art robustness and accuracy.

\section{Conclusion}
\label{sec:conclusion}
\vspace{-0.05in}

This paper presents a hybrid rotation averaging method that is robust to outliers. We combine fast view graph 
filtering to increase graph sparsity with state-of-the-art implementations of both global and local 
optimization methods.  The exposition is rounded off by a soft embedding into an incremental 
SfM pipeline leading to accurate, reliable, and highly efficient results. However, our method solves the rotation 
averaging problem all in once, thus it can also meet memory limitation in larger scenes. In our future work,
we are interested in extending this work in larger scenes with a more efficient manner.

\noindent \textbf{Acknowledgement:} We sincerely thank Prof. Frank Dalleart and Jing Wu for the discussion of the experimental details, 
and also for their help in improving the experimental results. L. Kneip furthermore acknowledges 
the support of the Natural Science Foundation of Shanghai (grant number 19ZR1434000), and his affiliation 
with the Shanghai Engineering Research Center of Intelligent Vision and Imaging.

{\small
\bibliographystyle{ieee_fullname}
\bibliography{egbib}

\begin{thebibliography}{10}\itemsep=-1pt

\bibitem{ceres-solver}
Sameer Agarwal and Keir Mierle.
\newblock Ceres solver.
\newblock \url{http://ceres-solver.org}.

\bibitem{DBLP:conf/iccv/AgarwalSSSS09}
Sameer Agarwal, Noah Snavely, Ian Simon, Steven~M. Seitz, and Richard Szeliski.
\newblock Building rome in a day.
\newblock In {\em {IEEE} 12th International Conference on Computer Vision},
  pages 72--79, 2009.

\bibitem{DBLP:journals/corr/Boumal15}
Nicolas Boumal.
\newblock A riemannian low-rank method for optimization over semidefinite
  matrices with block-diagonal constraints.
\newblock {\em CoRR}, abs/1506.00575, 2015.

\bibitem{DBLP:journals/ftml/BoydPCPE11}
Stephen~P. Boyd, Neal Parikh, Eric Chu, Borja Peleato, and Jonathan Eckstein.
\newblock Distributed optimization and statistical learning via the alternating
  direction method of multipliers.
\newblock {\em Foundations and Trends in Machine Learning}, 3(1):1--122, 2011.

\bibitem{DBLP:books/cu/BV2014}
Stephen~P. Boyd and Lieven Vandenberghe.
\newblock {\em Convex Optimization}.
\newblock Cambridge University Press, 2014.

\bibitem{DBLP:journals/mp/BurerM03}
Samuel Burer and Renato D.~C. Monteiro.
\newblock A nonlinear programming algorithm for solving semidefinite programs
  via low-rank factorization.
\newblock {\em Math. Program.}, 95(2):329--357, 2003.

\bibitem{DBLP:conf/icra/BustosCER19}
{\'{A}}lvaro~Parra Bustos, Tat{-}Jun Chin, Anders~P. Eriksson, and Ian~D. Reid.
\newblock Visual {SLAM:} why bundle adjust?
\newblock In {\em International Conference on Robotics and Automation}, 2019.

\bibitem{DBLP:conf/iccv/ChatterjeeG13}
Avishek Chatterjee and Venu~Madhav Govindu.
\newblock Efficient and robust large-scale rotation averaging.
\newblock In {\em {IEEE} International Conference on Computer Vision}, pages
  521--528, 2013.

\bibitem{DBLP:journals/pami/ChatterjeeG18}
Avishek Chatterjee and Venu~Madhav Govindu.
\newblock Robust relative rotation averaging.
\newblock {\em {IEEE} Trans. Pattern Anal. Mach. Intell.}, 40(4):958--972,
  2018.

\bibitem{DBLP:journals/toms/ChenDHR08}
Yanqing Chen, Timothy~A. Davis, William~W. Hager, and Sivasankaran
  Rajamanickam.
\newblock Algorithm 887: Cholmod, supernodal sparse cholesky factorization and
  update/downdate.
\newblock {\em {ACM} Trans. Math. Softw.}, 35(3):22:1--22:14, 2008.

\bibitem{DBLP:conf/cvpr/CrandallOSH11}
David~J. Crandall, Andrew Owens, Noah Snavely, and Dan Huttenlocher.
\newblock Discrete-continuous optimization for large-scale structure from
  motion.
\newblock In {\em {IEEE} Conference on Computer Vision and Pattern
  Recognition}, pages 3001--3008, 2011.

\bibitem{DBLP:journals/pami/CrandallOSH13}
David~J. Crandall, Andrew Owens, Noah Snavely, and Daniel~P. Huttenlocher.
\newblock {SfM} with {MRFs}: Discrete-continuous optimization for large-scale
  structure from motion.
\newblock {\em {IEEE} Trans. Pattern Anal. Mach. Intell.}, 35(12):2841--2853,
  2013.

\bibitem{DBLP:conf/cvpr/CuiGSH17}
Hainan Cui, Xiang Gao, Shuhan Shen, and Zhanyi Hu.
\newblock Hsfm: Hybrid structure-from-motion.
\newblock In {\em {IEEE} Conference on Computer Vision and Pattern
  Recognition}, pages 2393--2402, 2017.

\bibitem{DBLP:conf/3dim/CuiSGH17}
Hainan Cui, Shuhan Shen, Xiang Gao, and Zhanyi Hu.
\newblock Batched incremental structure-from-motion.
\newblock In {\em International Conference on 3D Vision}, pages 205--214, 2017.

\bibitem{DBLP:conf/iccv/CuiT15}
Zhaopeng Cui and Ping Tan.
\newblock Global structure-from-motion by similarity averaging.
\newblock In {\em {IEEE} International Conference on Computer Vision}, pages
  864--872, 2015.

\bibitem{DBLP:conf/eccv/DellaertRWMC20}
Frank Dellaert, David~M. Rosen, Jing Wu, Robert Mahony, and Luca Carlone.
\newblock Shonan rotation averaging: Global optimality by surfing {$SO(p)^n$}.
\newblock In {\em European Conference on Computer Vision}, volume 12351, pages
  292--308. Springer, 2020.

\bibitem{EriksonOKC20}
Anders Eriksson, Carl Olsson, Fredrik Kahl, and Tat-Jun Chin.
\newblock Rotation averaging with the chordal distance: Global minimizers and
  strong duality.
\newblock {\em {IEEE} Trans. Pattern Anal. Mach. Intell.}, 2020.

\bibitem{DBLP:conf/cvpr/ErikssonOKC18}
Anders~P. Eriksson, Carl Olsson, Fredrik Kahl, and Tat{-}Jun Chin.
\newblock Rotation averaging and strong duality.
\newblock In {\em {IEEE} Conference on Computer Vision and Pattern
  Recognition}, pages 127--135, 2018.

\bibitem{DBLP:journals/cacm/FischlerB81}
Martin~A. Fischler and Robert~C. Bolles.
\newblock Random sample consensus: {A} paradigm for model fitting with
  applications to image analysis and automated cartography.
\newblock {\em Commun. {ACM}}, 24(6):381--395, 1981.

\bibitem{DBLP:conf/accv/FredrikssonO12}
Johan Fredriksson and Carl Olsson.
\newblock Simultaneous multiple rotation averaging using {L}agrangian duality.
\newblock In {\em Asian Conference on Computer Vision}, volume 7726, pages
  245--258, 2012.

\bibitem{DBLP:conf/eccv/GoldsteinHLVS16}
Thomas Goldstein, Paul Hand, Choongbum Lee, Vladislav Voroninski, and Stefano
  Soatto.
\newblock Shapefit and shapekick for robust, scalable structure from motion.
\newblock In {\em Computer Vision - {ECCV} 2016 - 14th European Conference},
  pages 289--304, 2016.

\bibitem{DBLP:conf/cvpr/Govindu01}
Venu~Madhav Govindu.
\newblock Combining two-view constraints for motion estimation.
\newblock In {\em {IEEE} Computer Society Conference on Computer Vision and
  Pattern Recognition}, pages 218--225, 2001.

\bibitem{DBLP:conf/cvpr/Govindu04}
Venu~Madhav Govindu.
\newblock Lie-algebraic averaging for globally consistent motion estimation.
\newblock In {\em {IEEE} Computer Society Conference on Computer Vision and
  Pattern Recognition}, pages 684--691, 2004.

\bibitem{DBLP:journals/ijcv/HartleyTDL13}
Richard~I. Hartley, Jochen Trumpf, Yuchao Dai, and Hongdong Li.
\newblock Rotation averaging.
\newblock {\em International Journal of Computer Vision}, 103(3):267--305,
  2013.

\bibitem{DBLP:conf/iccv/JiangCT13}
Nianjuan Jiang, Zhaopeng Cui, and Ping Tan.
\newblock A global linear method for camera pose registration.
\newblock In {\em {IEEE} International Conference on Computer Vision}, pages
  481--488, 2013.

\bibitem{DBLP:conf/cvpr/KneipSS11}
Laurent Kneip, Davide Scaramuzza, and Roland Siegwart.
\newblock A novel parametrization of the perspective-three-point problem for a
  direct computation of absolute camera position and orientation.
\newblock In {\em The 24th {IEEE} Conference on Computer Vision and Pattern
  Recognition}, pages 2969--2976, 2011.

\bibitem{DBLP:journals/jscic/LaiO14}
Rongjie Lai and Stanley Osher.
\newblock A splitting method for orthogonality constrained problems.
\newblock {\em J. Sci. Comput.}, 58(2):431--449, 2014.

\bibitem{ma2012invitation}
Yi Ma, Stefano Soatto, Jana Ko{\v s}eck{\' a}, and S.~Shankar Sastry.
\newblock {\em An Invitation to {3-D} Vision: From Images to Geometric Models}.
\newblock Springer Science \& Business Media, 2012.

\bibitem{DBLP:conf/iccv/MoulonMM13}
Pierre Moulon, Pascal Monasse, and Renaud Marlet.
\newblock Global fusion of relative motions for robust, accurate and scalable
  structure from motion.
\newblock In {\em {IEEE} International Conference on Computer Vision}, pages
  3248--3255, 2013.

\bibitem{DBLP:books/sp/NocedalW99}
Jorge Nocedal and Stephen~J. Wright.
\newblock {\em Numerical Optimization}.
\newblock Springer, 1999.

\bibitem{DBLP:conf/cvpr/OzyesilS15}
Onur {\"{O}}zyesil and Amit Singer.
\newblock Robust camera location estimation by convex programming.
\newblock In {\em {IEEE} Conference on Computer Vision and Pattern
  Recognition}, pages 2674--2683, 2015.

\bibitem{park2017general}
Jaehyun Park and Stephen Boyd.
\newblock General heuristics for nonconvex quadratically constrained quadratic
  programming.
\newblock {\em arXiv:1703.07870}, 2017.

\bibitem{rosen2019se}
David~M Rosen, Luca Carlone, Afonso~S Bandeira, and John~J Leonard.
\newblock {SE-S}ync: A certifiably correct algorithm for synchronization over
  the special {E}uclidean group.
\newblock {\em International Journal of Robotics Research}, 38(2-3):95--125,
  2019.

\bibitem{DBLP:conf/cvpr/SchonbergerF16}
Johannes~L. Sch{\"{o}}nberger and Jan{-}Michael Frahm.
\newblock Structure-from-motion revisited.
\newblock In {\em 2016 {IEEE} Conference on Computer Vision and Pattern
  Recognition}, pages 4104--4113, 2016.

\bibitem{DBLP:conf/eccv/ShenZFZQ16}
Tianwei Shen, Siyu Zhu, Tian Fang, Runze Zhang, and Long Quan.
\newblock Graph-based consistent matching for structure-from-motion.
\newblock In {\em European Conference on Computer Vision}, volume 9907, pages
  139--155, 2016.

\bibitem{DBLP:conf/iccv/SweeneySHTP15}
Chris Sweeney, Torsten Sattler, Tobias H{\"{o}}llerer, Matthew Turk, and Marc
  Pollefeys.
\newblock Optimizing the viewing graph for structure-from-motion.
\newblock In {\em {IEEE} International Conference on Computer Vision}, pages
  801--809, 2015.

\bibitem{DBLP:journals/corr/abs-1903-00597}
Yulun Tian, Kasra Khosoussi, and Jonathan~P. How.
\newblock Block-coordinate minimization for large {SDP}s with block-diagonal
  constraints.
\newblock {\em arXiv: 1903.00597}, 2019.

\bibitem{DBLP:journals/corr/abs-1911-03721}
Yulun Tian, Kasra Khosoussi, David~M. Rosen, and Jonathan~P. How.
\newblock Distributed certifiably correct pose-graph optimization.
\newblock {\em arXiv: 1911.03721}, 2019.

\bibitem{DBLP:conf/iccvw/TriggsMHF99}
Bill Triggs, Philip~F. McLauchlan, Richard~I. Hartley, and Andrew~W.
  Fitzgibbon.
\newblock Bundle adjustment - {A} modern synthesis.
\newblock In {\em Vision Algorithms: Theory and Practice, International
  Workshop on Vision Algorithms}, pages 298--372, 1999.

\bibitem{DBLP:conf/cvpr/TronZD16}
Roberto Tron, Xiaowei Zhou, and Kostas Daniilidis.
\newblock A survey on rotation optimization in structure from motion.
\newblock In {\em {IEEE} Conference on Computer Vision and Pattern Recognition
  Workshops}, pages 1032--1040, 2016.

\bibitem{DBLP:journals/ini/WangS13}
Lanhui Wang and Amit Singer.
\newblock Exact and stable recovery of rotations for robust synchronization.
\newblock {\em Information and Inference}, 2(2):145--194, 2013.

\bibitem{DBLP:journals/corr/WangCK17}
Po{-}Wei Wang, Wei{-}Cheng Chang, and J.~Zico Kolter.
\newblock The mixing method: coordinate descent for low-rank semidefinite
  programming.
\newblock {\em arXiv: 1706.00476}, 2017.

\bibitem{Wen09rowby}
Zaiwen Wen, Donald Goldfarb, Shiqian Ma, and Katya Scheinberg.
\newblock Row by row methods for semidefinite programming.
\newblock Technical report, 2009.

\bibitem{DBLP:journals/mpc/WenGY10}
Zaiwen Wen, Donald Goldfarb, and Wotao Yin.
\newblock Alternating direction augmented lagrangian methods for semidefinite
  programming.
\newblock {\em Math. Program. Comput.}, 2(3-4):203--230, 2010.

\bibitem{DBLP:conf/eccv/WilsonS14}
Kyle Wilson and Noah Snavely.
\newblock Robust global translations with 1dsfm.
\newblock In {\em Computer Vision European Conference}, pages 61--75, 2014.

\bibitem{DBLP:conf/3dim/Wu13}
Changchang Wu.
\newblock Towards linear-time incremental structure from motion.
\newblock In {\em 2013 International Conference on 3D Vision}, pages 127--134,
  2013.

\bibitem{DBLP:conf/cvpr/ZachKP10}
Christopher Zach, Manfred Klopschitz, and Marc Pollefeys.
\newblock Disambiguating visual relations using loop constraints.
\newblock In {\em {IEEE} Conference on Computer Vision and Pattern
  Recognition}, pages 1426--1433, 2010.

\bibitem{DBLP:conf/cvpr/ZhangCL18}
Qianggong Zhang, Tat{-}Jun Chin, and Huu~Minh Le.
\newblock A fast resection-intersection method for the known rotation problem.
\newblock In {\em 2018 {IEEE} Conference on Computer Vision and Pattern
  Recognition}, pages 3012--3021, 2018.

\bibitem{zhu2017parallel}
Siyu Zhu, Tianwei Shen, Lei Zhou, Runze Zhang, Jinglu Wang, Tian Fang, and Long
  Quan.
\newblock Parallel structure from motion from local increment to global
  averaging.
\newblock {\em arXiv: 1702.08601}, 2017.

\bibitem{DBLP:conf/cvpr/ZhuZZSFTQ18}
Siyu Zhu, Runze Zhang, Lei Zhou, Tianwei Shen, Tian Fang, Ping Tan, and Long
  Quan.
\newblock Very large-scale global {SfM} by distributed motion averaging.
\newblock In {\em {IEEE} Conference on Computer Vision and Pattern
  Recognition}, pages 4568--4577, 2018.

\bibitem{DBLP:journals/cvpr/ZhuangCL18}
Bingbing Zhuang, Loong{-}Fah Cheong, and Gim~Hee Lee.
\newblock Baseline desensitizing in translation averaging.
\newblock In {\em {IEEE} Conference on Computer Vision and Pattern
  Recognition}, pages 4539--4547, 2018.

\end{thebibliography}
}

\newpage
\section*{APPENDIX}
\setcounter{equation}{0}
\setcounter{section}{0}
\setcounter{subsection}{0}

\section{Derivation of Globally Optimal Rotation Averaging}
Give a set of relative rotations $\{R_{ij}\}$, where $i, j \in [n]$, the aim of rotation averaging is to 
obtain the absolute rotations $\{R_i\}$ that satisfy the constraints
\begin{equation}
\label{equ:rotation_constraint}
    R_{ij} = R_jR_i^{-1} = R_j R_i^T
\end{equation}
between absolute and relative rotations. Usually there are more edges than nodes in an undirected graph, 
so there are more constraints than unknowns. Rotation averaging can be formulated as
\begin{equation}
  \label{equ:rotation_averaging_appendix}
  \min_{R_1, \cdots, R_n}\ \sum_{(i,j) \in E} d^p(R_{ij}, R_j R_i^T),
\end{equation}

Chordal distances is popular as the distance measure in~\eqref{equ:rotation_averaging_appendix}. 
Each residual along an edge of $\mathcal{G}$ will hence read
\begin{align}
\label{equ:rotation_averaging_frobenius}
  &\left\|R_j - R_{ij}R_i\right\|^2_F \\  \nonumber
  =& \left\|R_j\right\|_F^2 - 2\tr(R_j^TR_{ij}R_i) + \left\|R_i\right\|_F^2 \\ \nonumber
  =& \ 6 - 2\tr(R_j^TR_{ij}R_i).
\end{align}
The set of absolute rotations can be represented by
\begin{equation}
    \label{equ:R_row_representation}
    R = [R_1\ R_2\ \cdots\ R_n],
\end{equation}
and the graph $\mathcal{G}$ can be represented by
\begin{equation}
\label{equ:graph_laplacian}
G = \left[
\begin{array}{cccc}
0 & a_{12}R_{12} & \cdots & a_{1n}R_{1n}\\
a_{21}R_{21} & 0 & \cdots & a_{2n}R_{2n}\\
\vdots & \vdots & \ddots & \vdots\\
a_{n1}R_{n1} & a_{n2}R_{n2} & \cdots & 0
\end{array}
\right],
\end{equation}
where $a_{ij} = 1$ if the edge between views $i$ and $j$ exists, and $a_{ij} = 0$ otherwise. By 
combining Eqs.~\eqref{equ:rotation_averaging_frobenius}, \eqref{equ:R_row_representation} and 
\eqref{equ:graph_laplacian}, problem~\eqref{equ:rotation_averaging_appendix} can be rewritten as
\begin{equation}
    \begin{split}
        \min_R \ (6 - \tr(R^TGR)) \Leftrightarrow \min_R \ -\tr(R^TGR).
    \end{split}
\end{equation}

Eriksson \emph{et al.} \cite{EriksonOKC20} proves that there is no duality gap between the primal problem and the corresponding dual problem if the maximum residuals stay below a certain threshold. For readers who are interested in the duality 
gap and global optimality proof, we kindly refer them to \cite{EriksonOKC20} for more details.

\section{Fast View Graph Filtering}
\begin{figure}[t]
  \centering
  \subfigure[Initial View Graph]
  {
    \begin{minipage}{0.4\linewidth}
        \includegraphics[width=1\linewidth]{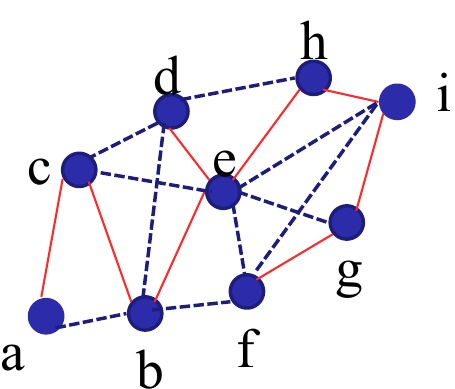}
    \end{minipage}
  }
  \subfigure[The first iteration]
  {
    \begin{minipage}{0.4\linewidth}
        \includegraphics[width=1\linewidth]{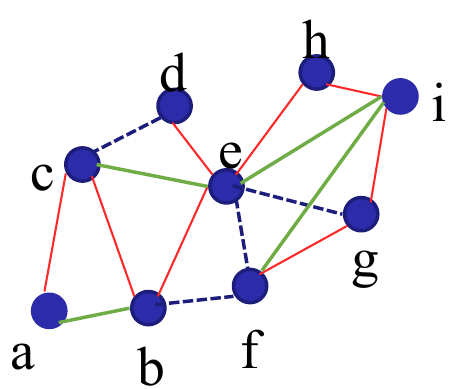}
    \end{minipage}
  }
  \subfigure[The second iteration]
  {
    \begin{minipage}{0.4\linewidth}
        \includegraphics[width=1\linewidth]{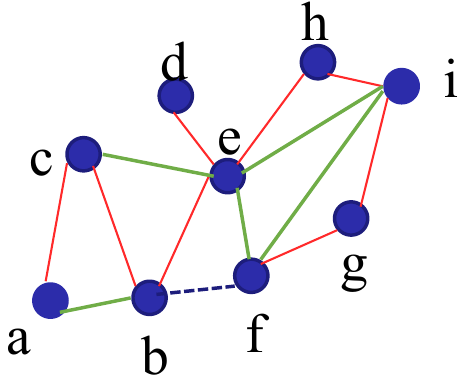}
    \end{minipage}
  }    
  \subfigure[The third iteration]
  {
    \begin{minipage}{0.4\linewidth}
      \includegraphics[width=1\linewidth]{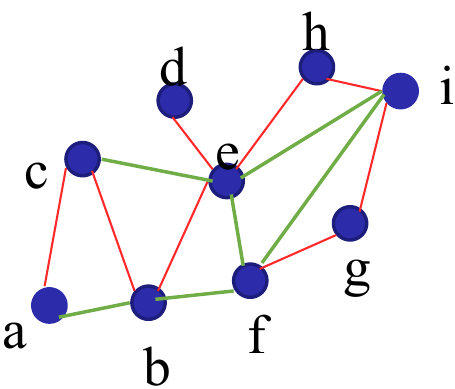}
    \end{minipage}
  }  
  \caption{The procedure of fast view graph filtering, where the blue nodes represent images,
           the \textit{red solid lines} represent the edges in the selected maximum spanning 
           tree (MaxST), the \textit{blue dotted lines} are edges that need to be validated,
           and the \textit{green solid lines} are edges that passed the verification.}
  \label{figure:fast_vgf}
\end{figure}

We present how the fast view graph filtering works in Fig.~\ref{figure:fast_vgf}.
In Fig.~\ref{figure:fast_vgf}, the blue nodes represent images,
the \textit{red solid lines} represent the edges in the selected maximum spanning tree (MaxST),
the \textit{blue dotted lines} are edges that need to be validated,
and the \textit{green solid lines} are edges that passed the verification.
In Fig.~\ref{figure:fast_vgf} (a), we deem the edges in the MaxST as valid. Then, we can collect
all the weak triplets based on the MaxST: $(a, c, b), (b, c, e), (e, d, b), (e, d, h), 
(g, f, i), (h, i, e)$ before the first iteration.

In Fig.~\ref{figure:fast_vgf} (b), edges $(a, b), (c, e), (e, i), (f, i)$ passed the verification,
edges $(d, b), (d, h)$ are deleted as they failed to pass the verification. Based on the first 
iteration, we also collect all the weak triplets $(c, e, d), (e, i, f), (e, i, g)$.
In Fig.~\ref{figure:fast_vgf} (c), $(c, d), (e, g)$ are deleted due to it fails to 
pass the verification, $(e, f)$ is marked by green solid line as it survives the validation.
Fig.~\ref{figure:fast_vgf} (d), we only need to validate weak triplet $(b, e, f)$,
suppose $(e, g)$ passed the verification and we mark it by green solid line.

\end{document}